\documentclass[10pt,twocolumn,letterpaper]{article}

\usepackage{wacv}
\usepackage{times}
\usepackage{epsfig}
\usepackage{graphicx}
\usepackage{amsmath}
\usepackage{amssymb}
\usepackage{url}
\usepackage{caption}
\usepackage{subcaption}
\usepackage[accsupp]{axessibility}
\graphicspath{ {./images/} }

\def\wacvPaperID{15} % Enter the WACV Paper ID here

\wacvfinalcopy % *** Uncomment this line for the final submission

\ifwacvfinal
\fi

\ifwacvfinal
\usepackage[breaklinks=true,bookmarks=false]{hyperref}
\else
\usepackage[pagebackref=true,breaklinks=true,colorlinks,bookmarks=false]{hyperref}
\fi

\ifwacvfinal
\fi

\begin{document}

%%%%%%%%% TITLE

% <- last ->

% \title{Real-time End-to-End License Plate Recognition Pipeline for Video Based Applications in Low Resource Deployment Scenarios}

\title{Real-time Bangla License Plate Recognition System for Low Resource Video-based Applications}

\author{
Alif Ashrafee\textsuperscript{1}\thanks{Both authors contributed equally to this research.}, Akib Mohammed Khan\textsuperscript{1}\footnotemark[1], 
Mohammad Sabik Irbaz\textsuperscript{2}, MD Abdullah Al Nasim\textsuperscript{2} \\
Department of Computer Science and Engineering, Islamic University of Technology\textsuperscript{1} \\
Machine Learning Team, Pioneer Alpha Ltd.\textsuperscript{2} \\
{\tt\small\{alifashrafee,akibmohammed,sabikirbaz\}@iut-dhaka.edu, nasim.abdullah@ieee.org}
% \textbf{A PREPRINT}
% \and
}

\maketitle
\thispagestyle{empty}

%%%%%%%%% ABSTRACT
\begin{abstract}
% introduce the need of the task
% our contributions and its importance
% final results
%   <- second last ->
\noindent Automatic License Plate Recognition systems aim to provide a solution for detecting, localizing, and recognizing license plate characters from vehicles appearing in video frames. However, deploying such systems in the real world requires real-time performance in low-resource environments. In our paper, we propose a two-stage detection pipeline paired with Vision API that provides real-time inference speed along with consistently accurate detection and recognition performance. We used a haar-cascade classifier as a filter on top of our backbone MobileNet SSDv2 detection model. This reduces inference time by only focusing on high confidence detections and using them for recognition. We also impose a temporal frame separation strategy to distinguish between multiple vehicle license plates in the same clip. Furthermore, there are no publicly available Bangla license plate datasets, for which we created an image dataset and a video dataset containing license plates in the wild. We trained our models on the image dataset and achieved an $AP_{0.5}$ score of 86\% and tested our pipeline on the video dataset and observed reasonable detection and recognition performance (82.7\% detection rate, and 60.8\% OCR F1 score) with real-time processing speed (27.2 frames per second).

\end{abstract}

% TODOs
% Final Checking
% Submission

%%%%%%%%% BODY TEXT
\section{Introduction}
\begin{figure}[h]
    \centering
    \includegraphics[width=\linewidth]{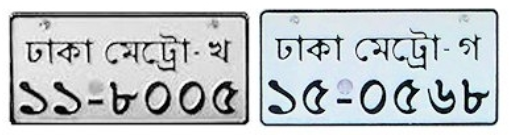}
    \caption{Example of Bangla License Plates. Source: \cite{plates}}
    \label{BDLP}
\end{figure}

\noindent In recent years there have been many advancements in developing Automatic License Plate Recognition (ALPR) systems and their necessity is apparent. In Bangladesh, there are over 4.47 million registered vehicles \cite{vehicles}. Yet, parking lot facilities are inadequate, and tracking cars entering or leaving parking lots is done manually and is generally disorganized. Due to the large volume of vehicles in metropolitan areas, it is difficult to find a vacant parking slot due to the lack of optimized parking systems. Furthermore, cars have to wait in long queues while they are manually logged one by one. This process can be very time-consuming for all involved parties. Hence, the demand for automated parking management systems has skyrocketed. In a parking system, by registering the license plate number of a car while it is entering and exiting, we can automatically generate a parking ticket fee using the time interval. ALPR systems can also be used to automate numerous other real-life application scenarios. It can be used to track guest vehicles in a private parking area. We can also automate toll collection systems and gas station management systems using similar methods. Examples of Bangla License Plates according to the Bangladesh Road Transport Authority (BRTA) format can be seen in figure \ref{BDLP}.

% Put example Images

\noindent Given the large variety of applications scenarios, ALPR systems have to be robust in detecting license plates in various conditions (unclear characters, plate variations, occlusions, illumination changes) and accurately recognize the characters while fulfilling the key criteria of real-time inference speed. Furthermore, if there are temporally separate instances of different vehicles in the same video, the system has to be able to correctly differentiate between them and store them separately. In such a system, each frame of a video needs to be processed to detect license plates but this comes at the cost of slower inference speed. Our proposed method contains a two-stage detection module. The first stage is a haar-cascade classifier \cite{soo2014object} that works as a wakeup mechanism to quickly discard frames with no license plates. The second stage is a MobileNet SSDv2 \cite{chiu2020mobilenet} detection model. The detection module stores only the best 3 cropped images for a single vehicle. Our pipeline also takes into account any interval between the appearance of two different vehicles so it can identify and separately store the result of each individual vehicle. Even with all these features, we have achieved real-time inference speed with satisfactory detection and recognition performance.

\noindent In summary, our contribution includes the creation of two datasets and their corresponding annotations. The first is a Bangla license plate image dataset for training purposes and the second is a video dataset that includes license plates in the wild for evaluating our proposed architecture. To our knowledge, this is the first annotated dataset of this type. These datasets have been made publicly available \footnote{https://tinyurl.com/p7e8tszk}. We introduce a pipeline that includes a two-stage detection module with the following features:
\begin{itemize}
    \item Wake-up mechanism for reducing inference time
    \item Best frame selection strategy
    \item Temporally separate vehicle instance identification
\end{itemize}
This method is highly optimized for low resource server-side run-time environments, thus providing a versatile solution to ALPR system deployment in various applications.

\section{Related Works}
\subsection{Real Time Object Detection}
\noindent Object detection has been a core concept and a challenging problem in Computer Vision since the earlier days. Paul Viola et. al. \cite{viola2001robust} introduced cascade classifiers trained using AdaBoost to detect faces in real-time. With the breakthrough in deep learning technologies in recent times, many advancements have been made in this particular field. Many object detection frameworks \cite{redmon2016you,ren2015faster,Howard2017MobileNetsEC} use convolutional neural networks to detect objects with performance close to human accuracy. Fast YOLO \cite{shafiee2017fast} produce an optimized architecture evolving the YoloV2 network in order to provide accurate detections in real-time. MobileNet SSDv2 \cite{chiu2020mobilenet} reduces computational complexity of object detection using a lightweight architecture based on MobileNetV2 \cite{mobilenetv2} providing real-time object detection in embedded devices. Object detection in videos is even more challenging since each frame needs to be processed which increases processing time. Shengyu Lu et. al. \cite{lu2019real} achieved real-time performance in video by using image preprocessing to eliminate the influence of image background with Fast YOLO as their object detection model. Han et. al. \cite{han2016seq} use temporal information to improve weaker detections in the same clip by using higher scoring detections in nearby frames. This helps to detect objects consistently across frames. For license plate detection, Montazolli et. al. \cite{8097294} uses a cascaded Fast YOLO architecture to detect larger regions around the license plate and then extract the license plate from those patches. They achieved a high precision and recall rate on a Brazilian license plate dataset. Hsu et. al. \cite{8078493} used a modified YoloV2 model for in-the-wild license plate detection, attaining a high FPS rate on a high-end GPU. Xie et al. \cite{article} took into consideration rotated license plates and proposed a rotation angle prediction for multi-directional license plate detection. Laroca et. al. \cite{laroca2018robust, laroca2019efficient} use a two-stage detection module to first localize the vehicle and then extract the license plate region from each frame. To deal with temporal redundancy, \cite{laroca2018robust} takes the union of all frames belonging to the same vehicle using majority vote.

\begin{figure*}[h]
    \centering
    \includegraphics[width=12cm]{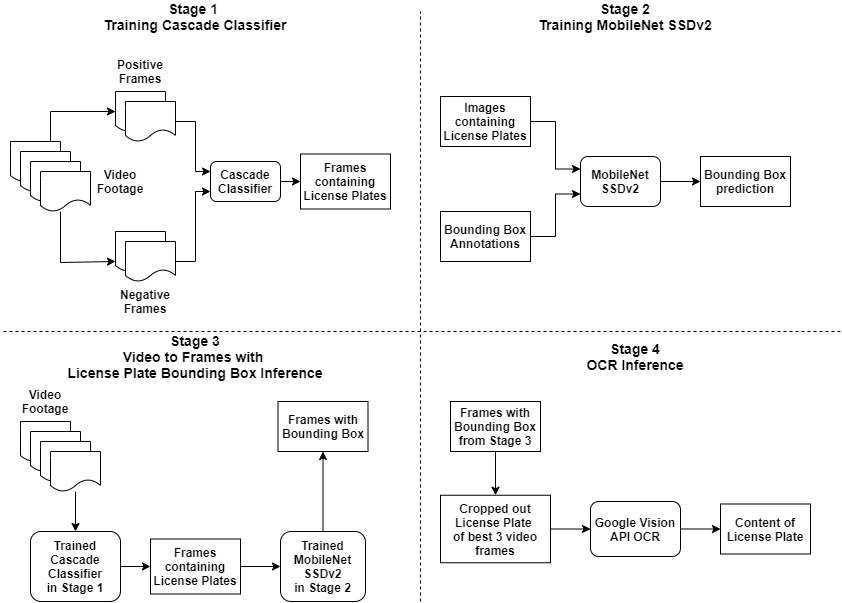}
    \caption{Proposed Methodology in 4 stages}
    \label{method}
\end{figure*}

\subsection{License Plate Recognition}
\noindent For an automatic license plate recognition system, the characters in the license plate need to be extracted and predicted accurately which is vital in many applications. Montazolli et. al. \cite{8097294} proposed a YOLO-based model to recognize the characters within a cropped license plate region, however, the recognition performance was poor with less than 65\% correctly recognized license plates. Li et. al. \cite{LI201814} performed character recognition without applying any segmentation. They modeled it as a sequence labeling problem where the sequential features were extracted using a CNN in a sliding window manner. To label the sequence they used a bi-directional recurrent neural network (BRNNs) with Long-Short-Term-Memory (LSTMs). Finally, they used a Connectionist Temporal Classification (CTC) network for sequence decoding. Zhuang et. al. \cite{zhuang2018towards} proposed a semantic segmentation with a character counting method to perform license plate character recognition. They used a simplified DeepLabV2 network for the semantic segmentation, performed Connected Component Analysis (CCA) for character generation, and used Inception-v3 and AlexNet for character classification and counting. In \cite{laroca2018robust}, Laroca et. al. used two networks to first segment the characters and then recognize them. In \cite{laroca2019efficient}, they recognize all the characters simultaneously by applying heuristic rules to adapt the results according to the predicted layout class. License plates can become blurred due to the fast motion of vehicles or when the extracted license plate patch is too small and needs to be enlarged. Many papers tackle this issue to boost recognition performance. Lu et. al. \cite{lu2016robust} deblur the license plate using a novel scheme based on sparse representation to identify the blur kernel. Svoboda et. al. \cite{svoboda2016cnn} implemented a text deblurring CNN to reconstruct blurred license plates. For Bangla license plates, character recognition is an even more complex issue due to its conjunct-consonants and grapheme roots \cite{chatterjee2019bengali}. Roy et al. \cite{roy2016license} proposed a boundary-based contour algorithm to detect the license plate in the vehicle region. Since the Bangla license plates contain information in two rows, they used horizontal projection with thresholding to separate the rows, and vertical projection with thresholding was used to separate the letters and numbers. Finally, they applied template matching for recognizing the characters. Dhar et. al. \cite{dhar2018system} introduced CNNs to extract features for character recognition. They also verified the shape of the license plate using the distance to borders (DtBs) method and corrected the horizontal tilting of plates using extrema points. Saif et. al. \cite{saif2019automatic} leverages the advantage of YOLO in that it can detect and localize objects simultaneously in an image using a single CNN. They fed the cropped license plate images after detection to a YOLOv3 model for segmenting and recognizing the characters and numbers. Their results, however, do not indicate real-time performance as they only achieved 9 frames per second on a Tesla K80 GPU while detecting and recognizing license plates in a video.

\section{Machine Learning Life Cycle}
\noindent There are currently no publicly available Bangla license plate datasets. Furthermore, to our knowledge, there are no active Bangla ALPR systems that perform real-time processing in video footage. So we had to work on all parts of the Machine Learning Life-cycle including Data Collection and Preparation, Machine Learning Modeling, and also a graphical user interface for Deployment and Interaction.

\subsection{Data Collection and Preparation}
\noindent There are many publicly available benchmark datasets for evaluating state-of-the-art (SOTA) ALPR systems. Current SOTA papers like Laroca et. al. \cite{laroca2018robust,laroca2019efficient} evaluate their methods on multiple datasets such as Caltech Cars, EnglishLP, UCSD-Stills, ChineseLP, AOLP, OpenALPR-EU, SSIG-SegPlate, and UFPR-ALPR. This kind of benchmark dataset is not available for Bangla License Plates. Hence, our main motivation in this paper was to create a public benchmark dataset for Bangla License Plates and evaluate our proposed method on that dataset to encourage further research on Bangla ALPR systems. To evaluate the performance of our model in real-time, we \textbf{separately} collected, annotated, and prepared a video dataset containing license plates in the wild. 

\begin{figure*}[h]
    \centering
    \includegraphics[width=11cm]{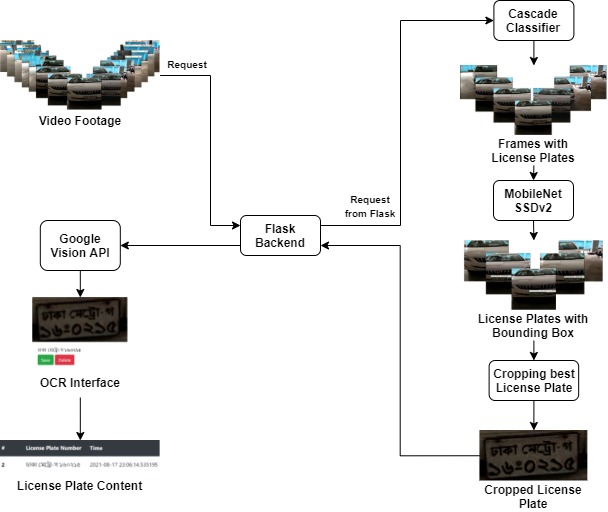}
    \caption{Overall Pipeline for Real-time Interaction}
    \label{pipeline}
\end{figure*}

\subsubsection{License Plate Image Dataset}
\noindent We created our own dataset of 1000 images by taking pictures of vehicles including their license plate from different distances, angles and lighting conditions around Dhaka City captured using varying models of smartphones. The images were of different sizes, later converted to 800 x 800 pixels and they are available in Joint Photographic Group (JPG) format. We manually annotated them by drawing bounding boxes around the license plate regions using a graphical image annotation tool Labellmg \cite{tzutalin} and finally split the whole dataset into an 80-20 percent train-validation split.

\subsubsection{License Plate Video Dataset}
\noindent We collected 79 video clips containing 98 license plates from different types of vehicles using crowd-sourcing, where each clip includes single or multiple vehicles from around different districts of Bangladesh. The videos were recorded using varying models of smartphones and are available in Moving Picture Experts Group-4 (MPEG-4) format having different sizes. We later converted all of the videos to a resolution of 480 x 480 pixels. Each video contains an average of 254 frames and the frame rate of each video was 24 frames per second (FPS). The videos were taken in a variety of locations, such as inside garages, car parks, lanes, and highways across the whole country. The license plates were of predominantly black text over white background while a few included black text over a green background. The authorized characters allowed on a license plate are listed and described by Ahsan et. al. \cite{ahsan2021intelligent}. Bangla license plates according to the BRTA format are composed of two lines where the first line contains the district name and the word 'Metro' followed by a hyphen and a character. The second line contains six Bangla digits that act as a unique identifier for the vehicle. Since videos were taken from several different districts, there is a good distribution of the authorized characters in our dataset. They were taken in the wild, with plate variations, artifacts on plates, occlusions, illumination changes, and unclear characters. We annotated the content of each plate: the ground truth of each character present in the plate, and the number of license plates present in each video to evaluate our entire model performance and speed from detection to recognition.

\subsection{Machine Learning Modeling}

\noindent We primarily focus on achieving real-time inference for each frame while also retaining satisfactory performance. As mentioned, feeding each frame to a large model is a drawback for our case because larger models have slower inference speeds. To tackle this issue, we first trained a lightweight haar-cascade classifier using video footage. Haar-cascade classifiers are trained using positive and negative examples \cite{opencv}. We extracted frames from the footage and labeled the frames accordingly for training. Frames with a license plate in them are labeled positive and frames containing only backgrounds are considered negative samples. A cascade classifier consists of multiple stages, where each stage is an ensemble of weak classifiers. Each stage is trained using boosting techniques that provide the ability to train a highly accurate classifier by taking a weighted average of the weak classifier results \cite{viola2001rapid}. Cascade classifiers divide the image into sub-windows. In the early stages, the weak classifiers are used to discard the negative sub-windows as fast as possible and detect all the positive instances to achieve low false-positive rates \cite{matlab}. Since the vast majority of windows do not contain license plates, this provides a very fast solution to the problem of inference time for each frame. Conversely, true positives are rare and worth taking the time to verify. Though this type of two-stage detection method was previously used \cite{laroca2018robust,laroca2019efficient}, the existing methods localize the vehicle in the first stage and then the number plate in the second stage. Our two-stage detector takes a different strategy. Instead of detecting the vehicle, the cascade classifier only tries to detect the license plate in the first stage which reduces the extra step of localizing the vehicle. This first stage is only used as a wake-up mechanism. The second stage is actually responsible for localizing the license plates.

\noindent For the second stage of our architecture, we use a more robust object detection model MobileNet SSDv2, which is a single shot detector. To train this model, we fed our custom dataset of license plate images along with their corresponding annotations to it. In the overall pipeline, the cascade classifier acts as a \textbf{wakeup mechanism} which is the primary filter for determining the presence of a license plate and if any license plate is detected by the cascade in any frame, that frame is then fed to the MobileNet SSDv2 module which acts as a second filter. MobileNet can draw more accurate bounding boxes \cite{deepa2019comparison} which we then use to localize the position of the license plate.

\noindent In our third stage, we handle multiple instances of license plates in the same video. We store each cropped instance of a detected license plate along with its corresponding confidence value in a dictionary. Since one vehicle appears in multiple consecutive frames, if we store each frame, there will be many redundant frames for each vehicle which is not preferable. In scenarios like parking or tolling systems, we consider all adjacent detected frames to belong to the same vehicle as long as the frames occur within a 24 frame or 1-second interval considering the videos are 24fps. Then we look into the dictionary and based on the highest confidence values, we store the best 3 cropped instances of a license plate. We follow this process for each following vehicle in the same video. Our approach is different from Laroca et al. \cite{laroca2018robust} They use temporal redundancy information to merge all the frames that belong to the same vehicle and extract the final output based on the most frequently predicted character at each LP position using a majority vote. We also provide users the ability to be able to choose between the 3 best frames for a particular vehicle so that they can store the most appropriate one for Optical Character Recognition (OCR).

\noindent For the final stage, the detected license plates are retrieved from the storage directory, enlarged, and passed to the Vision API \cite{vision} recognition module that returns the content of the license plate in string format. Some ALPR systems tend to propose end-to-end solutions. However, we kept the character recognition part separate to integrate human supervision which yields better results than an end-to-end pipeline. This is because Bangla License Plates (LPs) recognition is a challenging part in itself. Bangla LPs have two rows of characters \ref{BDLP}. As mentioned before, having two lines with the district identifier, followed by "Metro", a single Bangla character, and then 6 more characters, make the whole license plate number a very long string to recognize. As a result, we have found through extensive testing that character segmentation \cite{laroca2018robust} and template matching algorithms do not perform well in recognizing all the characters.

\begin{table*}[]
\begin{tabular}{|c|c|c|c|c|c|}
\hline
\textbf{Pipeline} & \textbf{Precision (\%)} & \textbf{Recall (\%)} & \textbf{F1 Score (\%)} & \textbf{Detection Rate (\%)} & \textbf{FPS} \\ \hline
YoloV3 Tiny & 60.2 & 55.6 & 57.1 & 75.5 & 11.1 \\
Cascade + YoloV3 Tiny & 56.3 & 51.6 & 53.1 & 69.4 & 27.1 \\
SSDv2 & \textbf{66.1} & 58.4 & 60.5 & \textbf{98.0} & 17.7 \\
Cascade + SSDv2 & 63.6 & \textbf{59.3} & \textbf{60.8} & 82.7 & \textbf{27.2} \\
\hline
\end{tabular}
\centering
\caption{Comparative result analysis of different pipelines}
\label{table1}
\end{table*}

\noindent We can summarize our proposed architecture in the following four stages as shown in figure \ref{method}.
\begin{itemize}
    \item \noindent \textbf{Stage 1}: We trained a cascade object detection model using positive and negative frames from video footage.
    \item \textbf{Stage 2}: Secondly, we use an object detection module MobileNet SSDv2. We feed our custom dataset of license plate images along with their corresponding annotations to the module in order to train it.
    \item \textbf{Stage 3}: We used our custom video dataset and fed it to the system where the frame in which a license plate exists, gets first detected by the cascade detector, and only then our MobileNet module detects it. The cascade detector acts as a wakeup mechanism for the MobileNet in order to increase efficiency. In the case of multiple instances of cars, we only accept that two license plates are different when there is a 24 or more frame gap between two detected license plates.
    \item \textbf{Stage 4:} The detected license plate is then cropped out, enlarged, and passed to the Vision API recognition module that returns the content of the license plate in string format.
\end{itemize}

\subsection{Deployment and Interaction}

\noindent We developed the whole system with user interaction and ease of use in mind. Since we aim to make ALPR systems more accessible, we built the system as a web app using Flask \cite{grinberg2018flask} framework so that it can be easily deployed. The overall pipeline of our system has been demonstrated in figure \ref{pipeline}.

\noindent On the left-hand side of the figure, we can see the interactive GUI components of the system. It starts with an interface where the user can upload video footage. Upon receiving a video as request, the flask framework sends it to the backend which can be seen on the right-hand side of the figure. On the backend side, at first, each frame is processed by the cascade classifier that separates only the frames containing license plates from the footage and feeds them forward to the backbone detection model, MobileNet SSDv2. It then detects and draws bounding boxes around the license plates. All of this processing is done in real-time and visual feedback of the detections is shown to the user on-screen. For each temporally separate vehicle appearing in the video, MobileNet SSDv2 stores 3 best-cropped license plate instances in the output directory. On the front-end side, the user can now view the detected plates. While fetching the stored license plates, Flask also sends a request to the Vision API to fetch the OCR result of each plate. Finally, the cropped license plates along with their corresponding OCR outputs are shown to the user. We also provide users the option for saving the OCR outputs in a database or deleting a particular result.

\section{Experimental Analysis}

%insert table here
\subsection{Experimental Setup}

\noindent A prime focus of our proposed method was to ensure real-time inference speed using minimal hardware specifications. For example, Laroca et al. \cite{laroca2018robust,laroca2019efficient} focus on high-speed inference using high-end GPUs like NVIDIA Titan XP. We had to consider the fact that the majority of the population in Bangladesh does not have access to such high-end GPUs. We thus ran all our inferences on a single-threaded CPU. Consequently, in table \ref{table1} we can see that our architecture can be deployed in real-time use case scenarios using very low resources.

\noindent At first we trained the MobileNet SSDv2 \cite{chiu2020mobilenet} model using a Tesla T4 GPU on Google Colaboratory. We set the batch size to 24, image size 300*300, intersection over union (IOU) threshold 0.35, score threshold $10^{-8}$, learning rate 0.004, momentum 0.9, decay 0.9 and RMSProp optimizer\cite{hinton2012neural}. We ran the training for 50 epochs.

\noindent For testing, all the experiments were conducted on a single thread of an Intel Core i5-7200U mobile \textbf{CPU} with 8GB of RAM. Each of the test videos was resized into 480*480 pixels to maintain consistency across the inference times. The yolov3 tiny model takes a blob of size 320*320 as input with a scale factor of 0.00392 to scale the values, mean of (0, 0, 0) which is subtracted from the values and swapRB set to True in order to swap the red and blue channels. Non-max suppression (NMS) is applied to the bounding boxes detected from the blob with a score threshold of 0.1 and NMS threshold of 0.4 which is also known as the IoU threshold. Finally, another thresholding is performed so that only the frames that have confidence higher than 0.1 are stored in the dictionary for a particular vehicle. Among all the frames stored in the dictionary for a single vehicle, 3 frames with the topmost confidence values are saved in the output directory in the end.

\noindent In the case of MobileNet SSDv2, a blob of 300*300 is extracted from each frame to feed into the network. Here, we only set the swapRB parameter to True. From the acquired predictions, we only take boxes with confidence values higher than 0.5 to be stored in the dictionary. Finally, the best 3 predictions are saved as output based on their confidence values.

\noindent For the cascade classifier, we give the entire 300* 300-pixel frame as input along with a scale factor of 1.1 which specifies how much the image size is reduced at each image scale, min neighbors set to 10 which determines how many neighbors each candidate rectangle should have to retain it and finally the min size set to 45*45 which is the lowest allowed size for a detection.

\begin{figure*}[]
    \centering
    \begin{subfigure}[b]{0.24\textwidth}
        \centering
        \includegraphics[width=3cm,height=3cm]{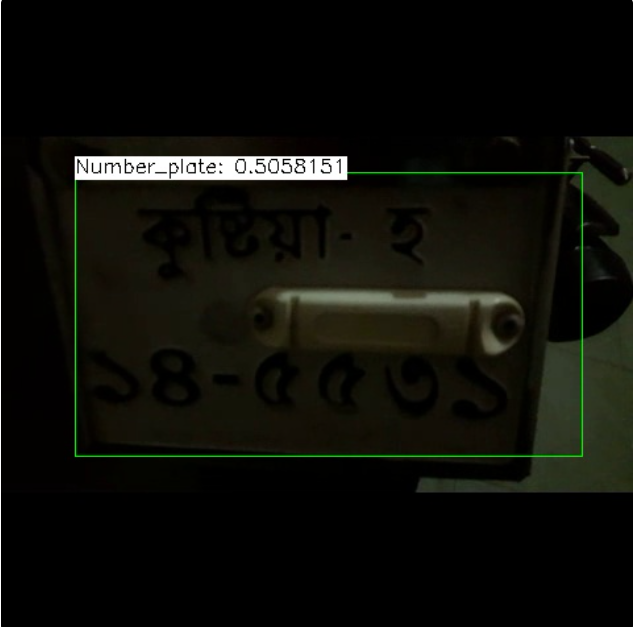}
    \end{subfigure}
    \begin{subfigure}[b]{0.24\textwidth}
        \centering
        \includegraphics[width=3cm,height=3cm]{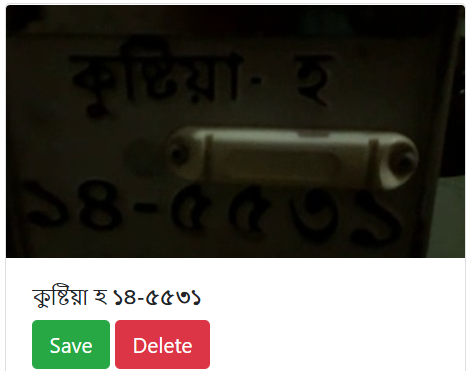}
    \end{subfigure}
    \begin{subfigure}[b]{0.24\textwidth}
        \centering
        \includegraphics[width=3cm,height=3cm]{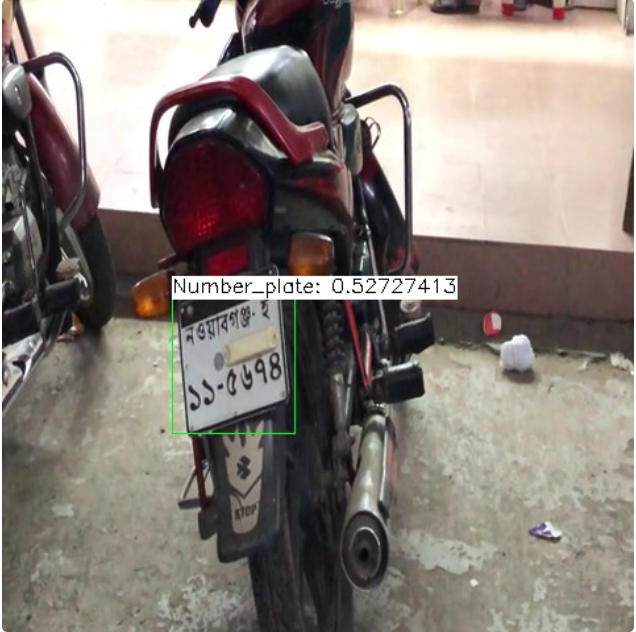}
    \end{subfigure}
    \begin{subfigure}[b]{0.24\textwidth}
        \centering
        \includegraphics[width=3cm,height=3cm]{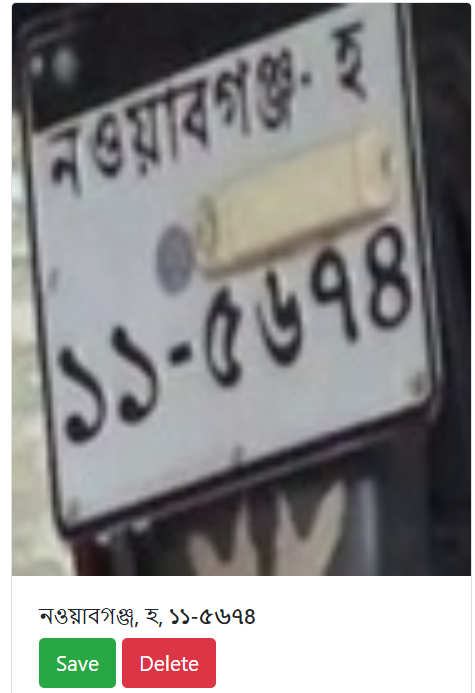}
    \end{subfigure}
\caption{License plates that are missed by Cascade + SSDv2 but are successfully detected by standalone SSDv2}
\label{fig:example1}
\end{figure*}

\begin{figure*}[]
    \centering
    \begin{subfigure}[b]{0.24\textwidth}
        \centering
        \includegraphics[width=3cm,height=3cm]{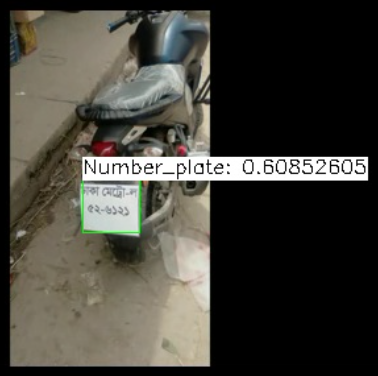}
    \end{subfigure}
    \begin{subfigure}[b]{0.24\textwidth}
        \centering
        \includegraphics[width=3cm,height=3cm]{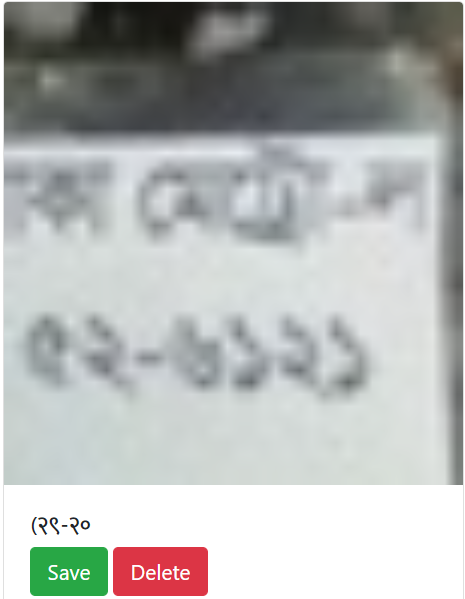}
    \end{subfigure}
    \begin{subfigure}[b]{0.24\textwidth}
        \centering
        \includegraphics[width=3cm,height=3cm]{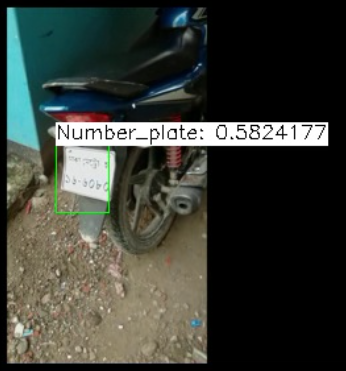}
    \end{subfigure}
    \begin{subfigure}[b]{0.24\textwidth}
        \centering
        \includegraphics[width=3cm,height=3cm]{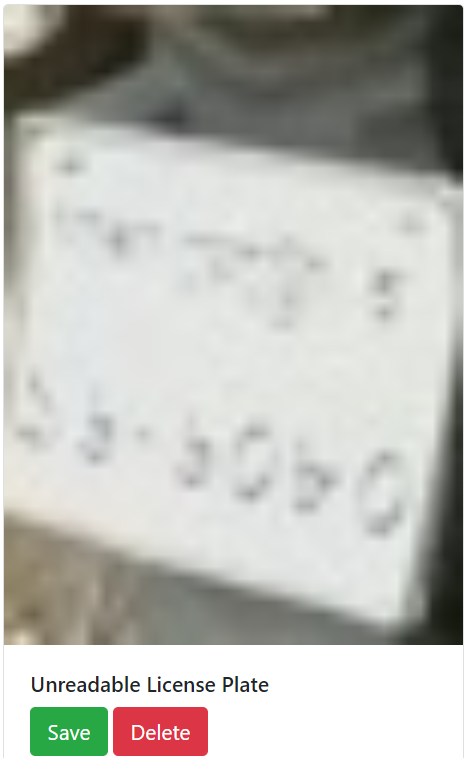}
    \end{subfigure}
\caption{License plates missed by Cascade + SSDv2 but detected by standalone SSDv2 that are not favorable for Vision API}
\label{fig:example2}
\end{figure*}

\subsection{Ablation Study}

\noindent We conducted 4 experiments to determine which pipeline works best for our task. The pipeline configurations and their corresponding results when applied on our license plate video dataset are outlined in Table \ref{table1}. Finally, we observe the obtained results and determine how and why a particular component leads to that particular result.

\noindent YoloV3 \cite{redmon2018yolov3} is an object detection framework that produces state-of-the-art detection results on the COCO dataset \cite{lin2014microsoft}. However, since our goal is to provide a real-time solution with minimal hardware resources, we used a more lightweight variant \textbf{YoloV3 tiny} model which decreases the depth of convolutional layers increasing the running speed significantly compared to the original YoloV3 network \cite{adarsh2020yolo}. This works as our backbone detection model and we consider this to be our baseline.

\noindent For the second pipeline, we used the same YoloV3 tiny model as our backbone detection model but with an additional \textbf{cascade classifier} as a wake-up mechanism. This is because YoloV3 tiny has much slower FPS rates which are not capable of real-time inference. Using the cascade classifier as a filter helps to quickly discard frames with no license plates. When applying only the cascade classifier on the license plate video dataset, we recorded an average of \textbf{41 FPS}. As a result, from Table \ref{table1} we can observe that this approach greatly reduces computational complexity and achieves a very high \textbf{FPS (27.1)} which is more than a \textbf{100\% gain} from the baseline FPS (11.1). Although this modification achieves real-time performance, it still does not have satisfactory detection rates or OCR accuracy.

\noindent In the aim of attaining greater detection and recognition performance, for our third pipeline, we used a trained \textbf{MobileNet SSDv2} as the main backbone detection model instead of YoloV3 tiny. Since there is a trade-off between performance and speed in detection models, MobileNet is an optimal choice for the backbone as discussed in \cite{deepa2019comparison,adarsh2020yolo}. We also remove the cascade component since we found from our experiments that it fails to identify license plates that are far away in the videos. For the standalone MobileNet however, this is not the case as it can detect distant plates easily. Consequently, from Table \ref{table1} we can observe a drastic improvement in the \textbf{detection rate (98.0\%)}. Nevertheless, in contrast to the detection rate, the F1 score (60.5\%) does not improve as substantially as compared to that of the baseline (57.1\%) and the second pipeline (53.1\%). This is because standalone MobileNet SSDv2 successfully detects a lot of plates that are small and far away but Vision API fails to make accurate predictions for such Bangla license plates. This experiment shows that using MobileNet SSDv2 as the backbone component instead of Yolov3 tiny has boosted both detection and recognition performance. While the speed of this pipeline (17.7 FPS) is slightly better than the baseline (11.1 FPS), real-time performance is still hindered as the average FPS drops by almost 10 FPS compared to the second pipeline (27.1) since MobileNet processes each and every frame much slower without the cascade component. Therefore, this pipeline is not adequate for real-time inference scenarios because we consider real-time videos to have at least \textbf{24 FPS}.

\noindent For our final pipeline, we kept the MobileNet SSDv2 as the backbone detection model and added the cascade component on top of it similar to the second pipeline. This yields in a much higher \textbf{FPS rate (27.2)} even with the MobileNet SSDv2 backbone while also gaining a higher \textbf{F1 score (60.8\%)} than the standalone MobileNet SSDv2 (60.5\%). Even though the detection rate of this pipeline (82.7\%) is much lower than that of the third pipeline (98.0\%), it still achieves a higher F1 score because the cascade component only feeds forward frames with clearly visible license plates to the MobileNet SSDv2 component that are in turn more favorable for the Vision API. As a result, with fewer plates detected than the third pipeline, this pipeline still provides more accurate OCR results than all the other pipelines because of the added layer of consistency that the cascade classifier provides.

\subsection{Result Analysis} 

\subsubsection{Evaluation Metrics}
The metrics used to calculate the recognition performance are determined by the Levenshtein distance \cite{levenshtein1965levenshtein} which measures the difference between two strings based on the number of insertions, deletions, and substitutions that have to be performed on the target string in order to match it with the reference string. According to the methods discussed in \cite{karpinski2018metrics}, we calculated the corresponding precision, recall, and F1 Score for recognition of each license plate. We applied pre-processing on both the ground truth and our OCR outputs such that they contain only the Bangla character set \cite{alam2020large} and numeric digits, converting the grapheme roots to individual consonants, and discarding any other character so that the Levenshtein distance is not affected by noise.

\subsubsection{Training Results}
We trained our MobileNet SSDv2 model on the license plate image dataset and obtained the results outlined in Table \ref{table2}. We achieved an $AP_{0.5}$ score of 86.2\%.
% insert example

\begin{table}[h]
\begin{tabular}{|c|c|c|c|c|c|c|}
\hline
\textbf{Model} & \textbf{AP} & \textbf{AP50} & \textbf{AP75} & \textbf{APs} & \textbf{APm} & \textbf{APl} \\ \hline
MobileNet & 0.47 & 0.86 & 0.47 & 0.13 & 0.46 & 0.55 \\ \hline
\end{tabular}
\centering
\caption{Training results on the License Plate image dataset}
\label{table2}
\end{table}

\begin{figure*}[]
\centering
\begin{subfigure}[b]{0.49\textwidth}
\centering
\includegraphics[height=2cm, width=6cm]{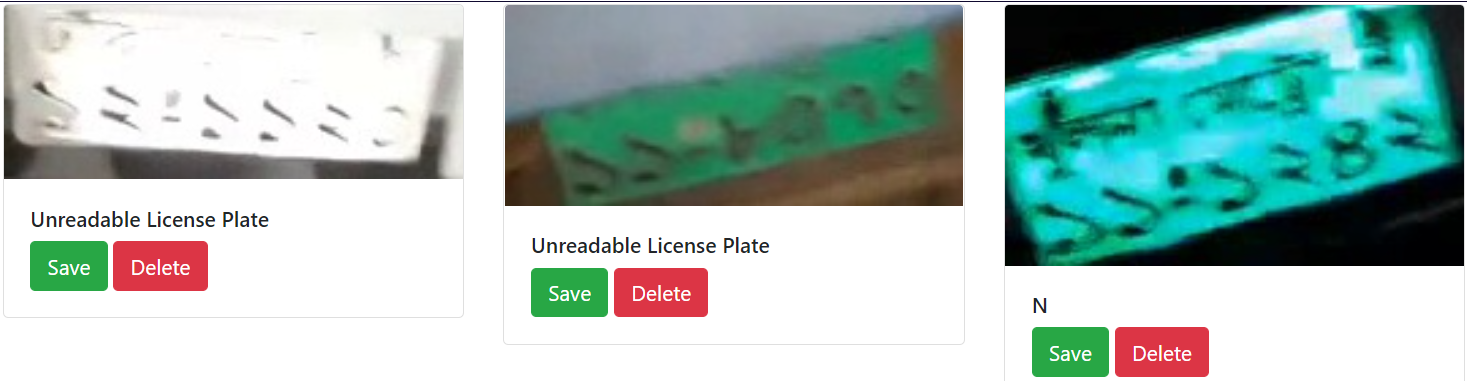}
\label{fig:subim1}
\end{subfigure}
\begin{subfigure}[b]{0.49\textwidth}
\centering
\includegraphics[height=2cm, width=6cm]{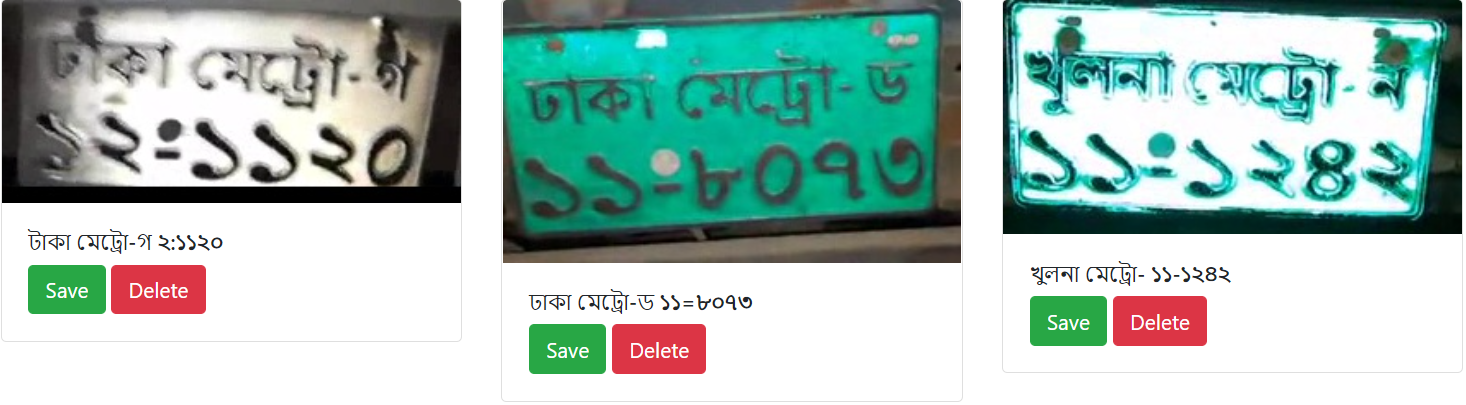}
\label{fig:subim2}
\end{subfigure}
\begin{subfigure}[b]{0.49\textwidth}
\centering
\includegraphics[height=2cm, width=6cm]{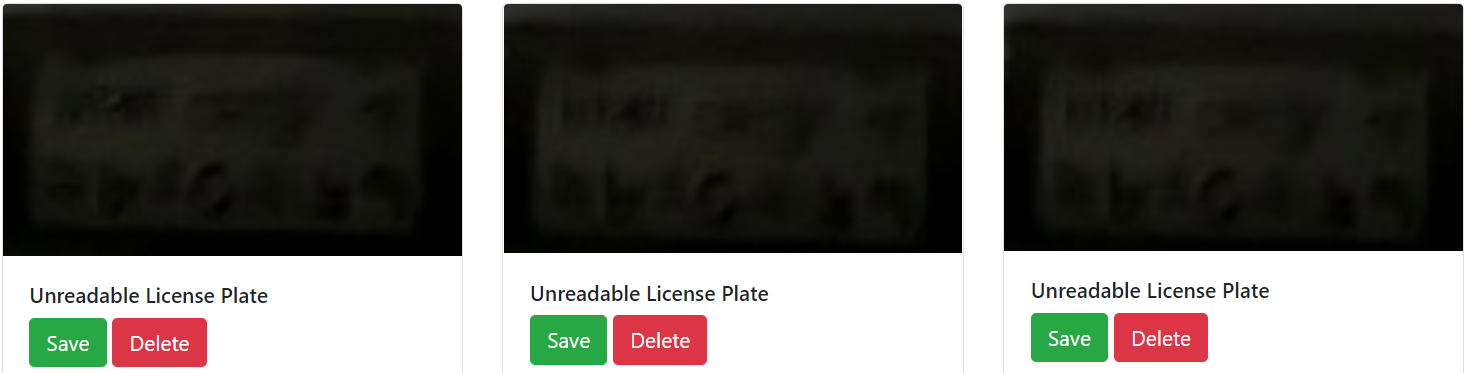}
\label{fig:subim3}
\end{subfigure}
\begin{subfigure}[b]{0.49\textwidth}
\centering
\includegraphics[height=2cm, width=6cm]{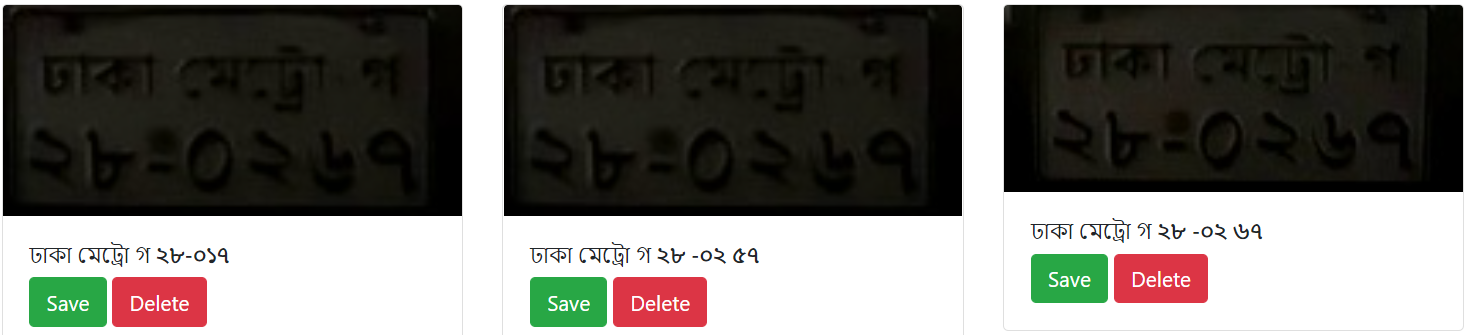}
\label{fig:subim1}
\end{subfigure}
\caption{OCR predictions from plates detected by standalone SSDv2 that are not favorable for Vision API are shown on the left, whereas detections by Cascade + SSDv2 that produce better OCR predictions are shown on the right}
\label{fig:example3}
\end{figure*}

%insert example

\subsubsection{Quantitative Results}
We evaluated our baseline and the other proposed pipelines on the license plate video dataset and outlined the results in Table \ref{table1}. From the ablation study, it can be observed that the SSDv2 and Cascade + SSDv2 pipelines have superior performance in all aspects in comparison to the other pipelines. Both of the pipelines have their respective merits and demerits. The detection accuracy for standalone SSDv2 is superior (98\%) at the cost of losing real-time performance while using the Cascade + SSDv2 results in much \textbf{faster FPS rate} (27.1 on average), but with a lower detection rate. The recognition performance of both pipelines using Vision API is similar (61\% F1 Score). The Bangla alphabet is made up of 11 vowels, 7 consonants, and 168 grapheme roots. This makes Bangla a very complex language having over 13,000 combinations compared to that of English which has only 250 \cite{chatterjee2019bengali}. As a result, license plates often contain a variety of conjunct-consonant characters which makes it difficult to perform OCR. Furthermore, we resized all videos to be 480*480 pixels which resulted in the extracted license plate patches being extremely small. When enlarged for OCR, these patches become blurred where Vision API fails to recognize most characters.

\subsubsection{Qualitative Results}
It is apparent from table \ref{table1} that the SSDv2 pipeline has a much higher detection rate than the Cascade + SSDv2 pipeline. In many scenarios, standalone SSDv2 can detect plates that are missed by the Cascade + SSDv2 pipeline since the cascade classifier is less sensitive to plates that are less visible, further away, or tilted. A few of these examples can be seen in figure \ref{fig:example1}. In a few of these cases, the frames extracted by standalone SSDv2 pipeline also yield highly accurate OCR predictions from Vision API as seen from the examples.

\noindent However, from our experiments we found that most of the time, the plates that are missed by Cascade + SSDv2 but detected by standalone SSDv2 pipeline are not very useful in terms of OCR prediction. This is because SSDv2 has a tendency of detecting plates that are further away with high confidence. These outputs are often either too small or too blurry and evidently unreadable. These detections certainly give the SSDv2 pipeline an edge over the Cascade + SSDv2 pipeline in terms of detection rate, but the OCR results show unsatisfactory results as seen from figure \ref{fig:example2}. We can also observe that the detections made by the Cascade + SSDv2 pipeline are much more reliable as cascade only detects plates that are clearly perceivable. This yields good OCR outputs as compared to the detections made by standalone SSDv2. We can see these detections and OCR predictions from figure \ref{fig:example3}.

\section{Conclusion and Future Works}
\noindent In this paper, we propose a pipeline that can reliably detect license plates from video footage. We developed a strategy to store only the 3 best-detected instances of a particular vehicle. Our pipeline also consists of a method for individually storing temporally separate instances of different vehicles appearing in the same video. As seen from the various experiments, each of the pipelines has its own merits and demerits. MobileNet SSDv2 provides a pipeline with a very high detection rate but at the cost of a slower inference speed. Whereas Cascade + SSDv2 has the fastest speed and the highest recognition accuracy among all the pipelines but has much lower detection rates. As our future work, we want to deploy our system in various real-life environments and verify its day-to-day performance. We also want to apply preprocessing mechanisms on the detected plates and build our own Bangla license plate OCR system that can improve the recognition performance compared to what the Vision API currently provides.

{\small
\bibliographystyle{ieee_fullname}
\bibliography{ref.bib}
}

\end{document}